# IMPROVED KNOWLEDGE DISTILLATION FOR LAND-USE IMAGE CLASSIFICATION

Arundhuti Sur, Abhiroop Chatterjee, Susmita Ghosh
Jadavpur University, India
{arundhuti02,abhiroopchat1998,susmitaghoshju}@gmail.com

Emmett Ientilucci
Rochester Institute of Technology, USA
ejipci@rit.edu

***Abstract*—In the present article, an improved Knowledge Distillation (KD) framework has been proposed for efficient compression of deep convolutional neural networks for land-use image classification task. Motivated by the need to achieve competitive classification accuracy while reducing computational complexity, a teacher-student learning paradigm is adopted in which a VGG16 network transfers knowledge to a lightweight MobileNetV2 model. The proposed framework integrates hard supervision from ground truth labels with a soft supervision strategy that combines Kullback–Leibler divergence and Cosine Similarity losses. Experiments conducted on three land-use datasets show that the proposed KD-based method yields improved performance, and achieves an accuracy of 99.04%, outperforming both baseline student training and single-loss distillation approaches, while retaining substantial model compression.**

## I. INTRODUCTION

Remote sensing image analysis has advanced significantly with the development of deep learning and high-resolution imagery. In contrast to earlier handcrafted methods [1], [2], recent deep learning techniques have enabled automatic feature learning and achieved significant performance gains [3]–[7]. Further research have emphasized deeper and more expressive architectures to enhance feature representation.

**Literature Study.** SAL-TS-Net, a two-stream densely connected CNN [8], enhances feature diversity, with an accuracy of 98.90% on the UCM dataset [9]. A DenseNet-based fully connected CNN [10] enables hierarchical feature extraction and improved gradient flow, producing an accuracy of 99.50% on the same dataset. A CNN–CapsNet framework [11] refines CNN features, giving an accuracy of 98.81%. Attention mechanisms further enhance CNNs by focusing on salient regions and discriminative channels. Moreover, Transformer-based architectures model long-range dependencies and global contextual relationships, achieving state-of-the-art performance. Further, CBAM+RDN, a residual dense CNN [12], integrates channel and spatial attention to highlight discriminative features yielding an accuracy of 99.82% on UCM, whereas, CSAT [13], a transformer architecture enhanced with channel–spatial attention, captures both local saliency and global context, giving an accuracy of 97.86%. UrbanScape-Net [14], a multi-scale CNN driven by spatial and self-attention extracts rich contextual features producing 98.83% accuracy. LG-ViT, a vision transformer jointly models local features and global context [15] through interactive learning, giving an accuracy of 99.93%. SwinHCST [16], a CNN–Transformer hybrid, gives 98.10%. Despite achieving high accuracy, advanced CNN and Transformer-based models incur high computational and memory costs.

**KD-based Approaches.** To address the above challenges, model compression techniques have gained prominence, with Knowledge Distillation (KD) [17] emerging as an effective solution. KD transfers knowledge from a high-capacity teacher to a lightweight student model, and enables strong performance with lesser parameters. In the context of remote sensing, KD has been successfully applied in several forms. Knowledge-distilled early-exit model [18] reduces inference cost and enables efficient deployment on edge platforms, with an accuracy of 97.9% on UCM. Relational and cluster-guided distillation methods [19], [20] further enhance student learning by preserving pairwise similarities, and improves accuracy over conventional KD approaches by producing 94.90% accuracy.

Most existing KD approaches in remote sensing rely on a single distillation objective and may fail to capture the complementary probabilistic, geometric, and structural knowledge encoded in deep models. Some recent studies, including Grassmann Manifold Knowledge Distillation [21] transfer geometric representations and attain 92.62% accuracy on UCM, while Multiscale Knowledge Distillation [22] that aligns features across network depths attaining accuracy of 98.22%.

Motivated by these observations, this article proposes a knowledge distillation framework that preserves semantic relationships and feature geometry to extract a high inter-class similarity in aerial imagery while maintaining computational efficiency. The proposed method effectively reduces model complexity and yields competitive classification performance.

## II. METHODOLOGY

Our framework follows a teacher–student learning paradigm. Initially, a larger teacher network is trained using supervised learning. The deep and uniform convolutional structure of VGG16 enables it to learn rich hierarchical features, and makes it an effective teacher for distillation. Subsequently, a smaller student network (MobileNetV2) is trained under the guidance of the teacher using a composite loss function combining hard classification loss with two soft distillation losses. Due to its reduced number of parameters while maintaining the capability of discriminative feature learning, MobileNetV2 functions as an effective student for this process. This strategy enables the student model to learn both discriminative class boundaries from ground truth and rich inter-class relationships encoded in the teacher's softened outputs.

### A. *Proposed Framework for Knowledge Distillation*

Knowledge distillation is a model compression technique where a large model (Teacher) transfers its learned knowledge to a smaller, faster model (Student) so that the student achieves high accuracy with fewer parameters. The teacher model provides not only the predicted class label but also a confidence distribution over all classes. Consequently, instead of training the student only on hard labels (ground truth), we also train it using the soft predictions of the teacher. By learning to match these probability distributions, the student captures both the target class information as well as the relative similarities among the classes. The block diagram of the proposed strategy is shown in Fig.1.

Knowledge distillation formulates student training as an optimization problem that minimizes two complementary objectives:

a) Hard loss- Standard classification loss with true labels measuring the discrepancy between student predictions and ground truth labels.
b) Soft loss- Measures the discrepancy between student and teacher outputs, indicating how effectively the student mimics the teacher.

Let y denote the ground truth label, $z_t$ and $z_s$ represent the logits produced by the teacher and student models, respectively, and $T$ be the temperature parameter used to soften the probability distributions. The softened output probabilities, $p_t$ for teacher and $p_s$ for student, are computed as:

$$p_t = \text{softmax}\left(\frac{z_t}{T}\right),\ p_s = \text{softmax}\left(\frac{z_s}{T}\right). \quad (1)$$

As stated, the temperature parameter controls the smoothness of the output distribution.

**Hard Loss-**

The hard loss, $\mathcal{L}_{hard}$, enforces correct classification with respect to true labels using the categorical cross-entropy loss:

$$\mathcal{L}_{hard} = -\sum_{c} y_c \log(p_{s,c}), \quad (2)$$

where $y_c$ denotes the ground truth label for class $c$, and $p_{s,c}$ is the predicted probability of the student s for class c. Minimizing this loss maximizes the log-likelihood of the correct class, encouraging the student network to assign high confidence to the true class while suppressing incorrect predictions. This enforces separability between class distributions in the output space and promotes discriminative feature learning.

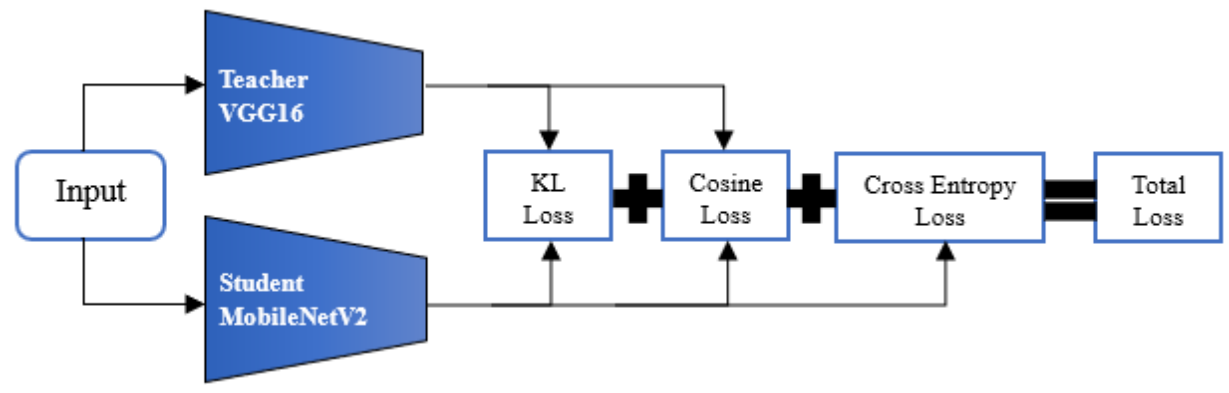

Fig. 1. Representation of the Proposed Knowledge Distillation Framework.

**Probabilistic-Representational Soft Loss-**

To overcome the limitations of distillation, a Probabilistic-Representational Soft Loss is introduced. By combining KL divergence and Cosine Similarity losses, our formulation allows the student network to learn both the conditional probability of the teacher model and the structural alignment of latent representations.

a) KL divergence Loss:

Temperature scaling is applied to the logits of both the teacher and student networks to soften the output probability distributions. The softened class probability for class $i$ is defined as follows.

The teacher probability distribution is written as:

$$p_t^{(i)} = \frac{exp(z_{t,i}/T)}{\sum_j exp(z_{t,j}/T)} \quad (3)$$

and the student probability distribution is given by:

$$p_s^{(i)} = \frac{exp(z_{s,i}/T)}{\sum_j exp(z_{s,j}/T)}, \quad (4)$$

where, $z_{t,i}$ and $z_{s,i}$ denote the logits produced by the teacher and the student network, respectively, for class $i$, $T$ controls the smoothness of the output distribution, $j$ indexes over all classes, and $p_t^{(i)}$ and $p_s^{(i)}$ represent the softened probability assigned to class $i$ by the teacher and the student model, respectively.

KL divergence computes the soft distillation loss, as it measures the divergence between the teacher and the student probability distributions. The KL divergence soft loss, $\mathcal{L}_{soft}^{\mathcal{KL}}$, is defined as:

$$\mathcal{L}_{soft}^{\mathcal{KL}} = T^2 \sum_i p_t^{(i)} \log\left(\frac{p_t^{(i)}}{p_s^{(i)}}\right). \quad (5)$$

Minimizing this loss encourages the student to approximate the teacher's output distribution, thereby transferring probabilistic knowledge in the form of relative class confidences.

b) Cosine Similarity Loss:

Logit normalization is applied to the outputs of both the teacher and the student networks prior to computing the Cosine Similarity loss, as follows:

$$\hat{z_t} = \frac{z_t}{||z_t||_2}, \quad \hat{z_s} = \frac{z_s}{||z_s||_2}, \tag{6}$$

where, $\hat{z_t}$ and $\hat{z_s}$ are the L2-normalized teacher and student logits, respectively; $z_t$ and $z_s$ are, respectively, the logit vectors produced by the teacher and student networks before normalization, and $||z||_2$ is L2 norm of the logit vector:

$$||z||_2 = \sqrt{\sum_i z_i^2}. \tag{7}$$

This normalization removes the influence of logit magnitude and preserves only the angular relationship between the two representations, ensuring that the distillation process focuses on structural alignment. Cosine Similarity is then employed to compute the soft distillation loss, as it effectively measures the directional agreement between the normalized teacher and student logits. The Cosine Similarity loss, $\mathcal{L}^{cos}$, is defined as:

$$\mathcal{L}^{cos} = 1 - \frac{\widehat{Z_t} \cdot \widehat{Z_s}}{||\widehat{Z_t}|| \, ||\widehat{Z_s}||}. \tag{8}$$

This captures representational similarity between teacher and student, thereby making it well-suited for aligning feature geometry during distillation. The said loss is in the range of [0, 2], where lower values indicate stronger alignment between the teacher and the student representations, facilitating effective transfer of representational knowledge.

c) Total Soft Loss-

The combined soft loss, $\mathcal{L}_{soft}$, is formulated as a weighted combination of the two loss components:

$$\mathcal{L}_{soft} = \lambda \mathcal{L}_{soft}^{KL} + (1 - \lambda)\mathcal{L}^{cos}, \tag{9}$$

where, $\lambda \in [0,1]$ controls the relative contributions of class-level semantic similarity and feature-level structural alignment.

**Total Distillation Loss-**

The final training objective, $\mathcal{L}_{total}$, for the student model is formulated as:

$$\mathcal{L}_{total} = \alpha \mathcal{L}_{hard} + (1 - \alpha)\mathcal{L}_{soft}. \tag{10}$$

The weighting parameter $\alpha \in [0,1]$ controls the trade-off between supervised learning and knowledge transfer.

### *B. Contributions*

The proposed methodology enables complementary knowledge transfer by integrating hard loss supervision, probabilistic distillation, and representational alignment. The hard loss (Eq. 2) preserves discriminative class boundaries by maintaining strong supervision from ground truth labels. The KL divergence soft loss (Eq. 5) transfers probabilistic knowledge by encoding relative class confidence and semantic similarity, guiding the student in assigning appropriate confidence to its predictions. Unlike conventional usage of Cosine Similarity as a fixed angular similarity metric in static feature space, the present work exploits Cosine Similarity as an additional component of the loss function (lower values indicated absolute alignment), which aims to mimic the directionality of the teacher (Eq. 8), by aligning the structural geometry of the teacher and the student representations. Together, these components provide improved performance.

Aerial imagery has high inter-class similarity and finer visual variations, which are effectively captured through this methodology. Preserving semantic relationships and feature geometry enables improved discrimination among visually similar classes and maintains computational efficiency.

## III. EXPERIMENTS AND RESULTS

As mentioned, experiments were conducted on the UC Merced land-use dataset [9]. It contains high-resolution aerial scene images for land-use/land-cover analysis in aerial imagery. There is a total of 2100 RGB images, each of size 256x256 pixels across diverse land-use categories like agricultural, airplane, forest, runway. It comprises 21 distinct land-use classes, and there are 100 images per class, ensuring a balanced class distribution. The spatial resolution is 0.3 m.

### *A. Image Preprocessing and Parameters*

Each RGB image was normalized by rescaling pixel values to the range [0, 1]. The train, validation, and test sets were split into 80:10:10. Batch size was set to 32. The proposed model is trained using AdamW [23] optimizer with a learning rate of $1\text{e}x10^{-5}$. Training was performed for a maximum of 100 epochs with early stopping based on the validation accuracy to prevent overfitting. Temperature, T, is set to 4.0. The temperature, T, and the weighting parameters $\alpha$ and $\lambda$ were tuned on the validation set by selecting values that yielded the best performance.

### *B. Performance Metrics*

To evaluate the performance of the proposed model, we have considered accuracy, precision, recall, and F1-score as classification metrics. In addition, computational efficiency was assessed in terms of FLOPs and the number of model parameters.

### *C. Results and Analysis*

Performance comparison of our KD-based approach with SOTA methods has been shown in Table I. It is seen that the distilled student model achieves maximum classification accuracy of 99.04% on the UCM dataset. The teacher network contains 14.85 million parameters with a computational cost of 20.06 GFLOPs, whereas the knowledge-distilled student model requires only 2.42 million parameters and 0.40 GFLOPs, resulting in a substantial reduction in computational overhead. The table shows that when compared to eight other non-KD-based baselines, our approach produces better accuracy for five cases. However, while comparing with recent

KD-based approaches, the proposed distillation strategy yields competitive performance.

TABLE I. ACCURACY COMPARISON WITH STATE-OF-THE-ART METHODS ON UC MERCED DATASET

| Type | Method | Accuracy (%) | Venues |
|---|---|---|---|
| Non-KD-Based | SAL-TS-Net [8] | 98.90 | RS '18 |
| | DenseNet-FCN [10] | 99.50 | MBE '19 |
| | VGG-16-CapsNet [11] | 98.81 | RS '19 |
| | CBAM+RDN [12] | 99.82 | RS '20 |
| | CSAT [13] | 97.86 | Sc. Reports '22 |
| | UrbanScape-Net [14] | 98.83 | IGARSS '24 |
| | LG-ViT [15] | 99.93 | GRSL '23 |
| | CNN+SWIN [16] | 98.10 | IJRS '23 |
| KD-Based | GFNet-KD-E3C [18] | 97.90 | ACM MM '25 |
| | CGKD [19] | 98.57 | JSTARS '25 |
| | PairKD [20] | 94.90 | RS '22 |
| | GMN-KD [21] | 92.62 | RS '21 |
| | MCKD [22] | 98.22 | TGRS '25 |
| | **OURS** | 99.04 | - |

In particular, compared to the recent MCKD [22], having ResNet34 as a teacher, our framework using VGG16 as a teacher and MobileNetV2 as a student, achieves higher classification accuracy than all of their student network configurations. While their approach gains accuracy improvement using deeper teachers ResNet50/ ResNet101 (25.6/44.5 million parameters) and ResNet34 as a student with 21.8 million parameters, our method achieves comparable accuracy with a much lightweight student (with 2.42 M parameters), offering a superior accuracy-parameter trade-off. The nearly identical accuracy, precision, recall, and F1-score, Table II show highly balanced classification performance with minimal class bias. Per-class accuracy analysis shows strong performance across classes with 100% accuracy except for dense residential at 80%. While the primary evaluation is done on the UC Merced dataset, preliminary experiments on the AID dataset [24] for 10 independent runs shows accuracy of 95.77±0.06% with highest being 95.85% and on the NWPU-RESISC45 dataset [25] for 10 independent runs shows accuracy of 91.59±0.11% with highest being 91.73%, demonstrating promising generalization. By combining hard supervision with KL-based probabilistic distillation and cosine similarity-based feature alignment, the enhanced loss function enables the lightweight student network to preserve discriminative capabilities.

TABLE II. CLASSIFICATION METRICS

| Accuracy (%) | Precision (%) | Recall (%) | F1-Score (%) |
|---|---|---|---|
| 99.04 | 99.11 | 99.04 | 99.04 |

### D. Ablation Study

An ablation study was conducted over 10 independent runs, to evaluate the impact of hard-label supervision, KL-divergence–based soft targets, and cosine similarity–based representational alignment, with statistical significance assessed using unpaired t-test. The same data split was used and evaluated on the test set for consistency. The baseline MobileNetV2 student achieves 96.47 ± 0.71% mean accuracy. Incorporating Hard + Cosine and Hard + KL distillation improves performance to 97.08 ± 0.61% and 97.17 ± 0.79%, respectively. Removing hard supervision and using only KL + Cosine soft distillation degrades performance to 95.14 ± 0.99%, thus highlighting the importance of hard labels. The full Hard + KL + Cosine distillation framework achieves the best results with 98.27 ± 0.74% mean accuracy and 99.04% peak accuracy, demonstrating the complementary benefits of all distillation components. With 18 degrees of freedom, the significance test yields a p-value of 3.05 $\times 10^{-5}$, indicating that the proposed model achieves a statistically significant performance improvement over the baseline, less than the widely used 0.05 threshold.

TABLE III. ABLATION STUDY OF KNOWLEDGE DISTILLATION LOSS COMPONENTS

| Model | Highest Accuracy | Mean Accuracy |
|---|---|---|
| VGG16-Teacher | 97.62% | 96.76 ± 0.70% |
| MobileNetV2-Student | 97.14% | 96.47 ± 0.71% |
| Hard + Cosine KD | 97.89% | 97.08 ± 0.61% |
| Hard + KL KD | 98.10% | 97.17 ± 0.79% |
| KL + Cosine KD | 96.19% | 95.14 ± 0.99% |
| **Hard + KL + Cosine KD** | **99.04%** | **98.27 ± 0.74%** |

## IV. CONCLUSION

The proposed Knowledge Distillation with enhanced loss function effectively transfers probabilistic and representational knowledge to a lightweight student network. The proposed approach achieves an overall accuracy of 99.04% on the UC Merced dataset while significantly reducing model size and computational complexity. These characteristics make the method well-suited for real-time and resource-constrained remote sensing applications. A potential limitation observed is some minor variance across simulations. While evaluation on the UC Merced dataset validates its effectiveness, extending the analysis to larger and more diverse aerial datasets with domain priors [26] will further strengthen its generalization capability. Future work will also focus on reducing model parameters and computational overhead on various tasks [27] while enhancing robustness through distillation strategies.

ACKNOWLEDGEMENT

This work received support from the IEEE GRSS under "ProjNET" scheme.

# REFERENCES

[1] T. Villmann, E. Merényi, and B. Hammer, “Neural Maps in Remote Sensing Image Analysis”, Neural Networks, vol. 16, pp. 389-403, 2003.

[2] M. Roy, S. Ghosh, and A. Ghosh, “A Neural Approach under Active Learning Mode for Change Detection in Remotely Sensed Images,” IEEE Journal of Selected Topics in Applied Earth Observations and Remote Sensing, vol. 7, pp. 1200-1206, 2013.

[3] G. Cheng, Z. Li, X. Yao, L. Guo, and Z. Wei, “Remote Sensing Image Scene Classification Using Bag of Convolutional Features,” IEEE Geoscience Remote Sensing Letters, vol. 14, pp. 1735–1739, 2017.

[4] S. Chen and Y. Tian, "Pyramid of Spatial Relations for Scene-Level Land Use Classification," IEEE Transactions on Geoscience and Remote Sensing, vol. 53, pp. 1947-1957, 2015.

[5] A. Chatterjee, S. Ghosh, and A. Ghosh, “Context-aware masking and learnable diffusion-guided patch refinement in transformers via sparse supervision for hyperspectral image classification,” In Proceedings of the IEEE/CVF International Conference on Computer Vision Workshops, pp. 2906-2915, 2025.

[6] Y. LeCun, Y. Bengio, and G. Hinton, “Deep Learning,” Nature, vol. 521, pp. 436-444, 2015.

[7] I. Goodfellow, Y. Bengio, and A. Courville, Deep Learning. MIT Press, 2016.

[8] Y. Yu and F. Liu “Dense Connectivity Based Two-Stream Deep Feature Fusion Framework for Aerial Scene Classification.” Remote Sensing, vol. 10, pp. 1158, 2018. https://doi.org/10.3390/rs10071158

[9] Y. Yang and S. Newsam, “Bag-Of-Visual-Words and Spatial Extensions for Land-Use Classification,” Proceedings 18th ACM SIGSPATIAL International Conference on Advances in Geographic Information Systems, pp. 270–279, 2010.

[10] J. Zhang, C. Lu, X. Li, H. Kim, and J. Wang, “A Full Convolutional Network Based on DenseNet for Remote Sensing Scene Classification,” Journal of Mathematical Biosciences and Engineering, vol. 16, pp. 3345-3367, 2019. doi: 10.3934/mbe.2019167

[11] W. Zhang, P. Tang, and L. Zhao “Remote Sensing Image Scene Classification Using CNN-CapsNet”. Remote Sensing, vol. 11, pp. 494, 2019.

[12] J. Zhao, J. Zhang, Z. Li, and J. Hwang, "Residual Dense Network Based on Channel-Spatial Attention for the Scene Classification of a High-Resolution Remote Sensing Image," Remote Sensing, vol. 12, pp. 1887, 2020. doi: 10.3390/rs12111887.

[13] J. Guo, N. Jia, and J. Bai, “Transformer Based on Channel-Spatial Attention for Accurate Classification of Scenes in Remote Sensing Image,” Scientific Reports, vol. 12, pp. 15473, 2022.

[14] A. Chatterjee, S. Ghosh, A. Ghosh, and E. Ientilucci, "Urbanscape-Net: A Spatial and Self-Attention Guided Deep Neural Network with Multi Scale Feature Extraction for Urban Land-Use Classification," IGARSS 2024 - 2024 IEEE International Geoscience and Remote Sensing Symposium, Greece, pp. 4884-4889, 2024.

[15] T. Peng, J. Yi, and Y. Fang, "A Local–Global Interactive Vision Transformer for Aerial Scene Classification," IEEE Geoscience and Remote Sensing Letters, vol. 20, pp. 1-5, 2023, Art no. 6004405.

[16] J. Song, Y. Fan, W. Song, H. Zhou, L. Yang, Q. Huang, Z. Jiang, C. Wang, and T. Liao, “SwinHCST: A Deep Learning Network Architecture for Scene Classification of Remote Sensing Images Based on Improved CNN and Transformer,” International Journal of Remote Sensing, vol. 44, pp. 7439-7463, 2023.

[17] G. Hinton, O. Vinyals, and J. Dean, "Distilling the Knowledge in a Neural Network," arXiv preprint arXiv:1503.02531, 2015.

[18] Y. Zhao, S. Li, and X. Feng, “Lightweight Remote Sensing Scene Classification on Edge Devices: An Early-Exit Distilled Global Filter Network Framework,” Proceedings 33rd ACM International Conference on Multimedia, pp. 11862–11870, 2025.

[19] Z. Deng, Z. Zhou, H. Zhao, X. Chen, D. Hong, and X. Sun, "Progressive Cluster-Guided Knowledge Distillation for Remote Sensing Image Scene Classification," IEEE Journal of Selected Topics in Applied Earth Observations and Remote Sensing, vol. 18, pp. 18068-18079, 2025. doi:10.1109/JSTARS.2025.3587845.

[20] H. Zhao, X. Sun, F. Gao, and J. Dong, “Pair-Wise Similarity Knowledge Distillation for RSI Scene Classification.” Remote Sensing, vol. 14, pp. 2483, 2022. https://doi.org/10.3390/rs14102483.

[21] L. Tian, Z. Wang, B. He, C. He, D. Wang, and D. Li “Knowledge Distillation of Grassmann Manifold Network for Remote Sensing Scene Classification.” Remote Sensing, vol. 13, pp. 4537, 2021. https://doi.org/10.3390/rs13224537.

[22] X. Li, L. Jiao, Q. Sun, F. Liu, X. Liu, L. Li, P. Chen, and S. Yang, "Multiscale Knowledge Distillation-Aware Classification for Remote Sensing Images," IEEE Transactions on Geoscience and Remote Sensing, vol. 63, pp. 1-15, 2025, Art no. 4509615. doi:10.1109/TGRS.2025.3604362.

[23] I. Loshchilov and F. Hutter, “Decoupled Weight Decay Regularization,” arXiv preprint, arXiv:1711.05101, 2017.

[24] G. S. Xia, J. Hu, F. Hu, B. Shi, X. Bai, Y. Zhong and L. Zhang, “AID: A Benchmark Dataset for Performance Evaluation of Aerial Scene Classification”, in IEEE Transactions on Geoscience and Remote Sensing, vol. 55, pp. 3965-3981, 2017. doi: 10.1109/TGRS.2017.2685945.

[25] G. Cheng, J. Han, X. Lu, “Remote Sensing Image Scene Classification: Benchmark and State of the Art”, arXiv preprint, https://doi.org/10.48550/arXiv.1703.00121, 2017

[26] A. Chatterjee and S. Ghosh “Learning Hyperspectral Images with Curated Text Prompts for Efficient Multimodal Alignment.”, arXiv preprint arXiv:2509.22697, 2025.

[27] S. Lekhak, E. Ientilucci, and A. Brinkley. "Viability of substituting handheld metal detectors with an airborne metal detection system for landmine and unexploded ordnance detection." *Remote Sensing* 16.24 (2024): 4732.